\documentclass[10pt]{article}
\usepackage[dvipsnames]{xcolor}

\usepackage{microtype}

\usepackage{url}

\usepackage{tikz}
\usepackage{ragged2e}

\usetikzlibrary{arrows, calc}

\usepackage{amsmath}
\usepackage{amssymb}
\usepackage{booktabs}

\usepackage[round]{natbib}
\usepackage{geometry}
\usepackage{graphicx}

\usepackage{charter}
\usepackage{hyperref}
\hypersetup{colorlinks=true,allcolors=BlueViolet}

\title{\textsc{\textbf{Earth System Data Cubes:\\Avenues for advancing Earth system research}}}

\author{
David Montero\textsuperscript{1,2,3,*}\quad
Guido Kraemer\textsuperscript{1,2}\quad
Anca Anghelea\textsuperscript{4}\quad
César Aybar\textsuperscript{5,6}\\
Gunnar Brandt\textsuperscript{7}\quad
Gustau Camps-Valls\textsuperscript{5}\quad
Felix Cremer\textsuperscript{8}\quad
Ida Flik\textsuperscript{1,2}\\
Fabian Gans\textsuperscript{8}\quad
Sarah Habershon\textsuperscript{1,2}\quad
Chaonan Ji\textsuperscript{1,2}\quad
Teja Kattenborn\textsuperscript{9}\\
Laura Martínez-Ferrer\textsuperscript{5}\quad
Francesco Martinuzzi\textsuperscript{1,2,10}\
Martin Reinhardt\textsuperscript{1,2,10}\\
Maximilian Söchting\textsuperscript{1,2,11}\quad
Khalil Teber\textsuperscript{1,2}\quad
Miguel D. Mahecha\textsuperscript{1,2,3,10,12}\\
\small\textsuperscript{1}RSC4Earth\quad \textsuperscript{2}IEF, Leipzig University\quad
\textsuperscript{3}iDiv\quad \textsuperscript{4}ESRIN, ESA\quad \textsuperscript{5}IPL, Universitat de Val\`encia\\
\small\textsuperscript{6}Water Competence Center\quad \textsuperscript{7}Brockmann Consult GmbH\quad
\textsuperscript{8}MPI for Biogeochemistry\\
\small\textsuperscript{9}Geosense, University of Freiburg\quad \textsuperscript{10}ScaDS.AI \quad \textsuperscript{11}ISP, Leipzig University\quad \textsuperscript{12}UFZ\\
\small*Corresponding author: \texttt{david.montero@uni-leipzig.de}
}
\date{}

\begin{document}

\maketitle

\begin{abstract}
Recent advancements in Earth system science have been marked by the exponential increase in the availability of diverse, multivariate datasets characterised by moderate to high spatio-temporal resolutions. 
Earth System Data Cubes (ESDCs) have emerged as one suitable solution for transforming this flood of data into a simple yet robust data structure. ESDCs achieve this by organising data into an analysis-ready format aligned with a spatio-temporal grid, facilitating user-friendly analysis and diminishing the need for extensive technical data processing knowledge.
Despite these significant benefits, the completion of the entire ESDC life cycle remains a challenging task.
Obstacles are not only of a technical nature but also relate to domain-specific problems in Earth system research.
There exist barriers to realising the full potential of data collections in light of novel cloud-based technologies, particularly in curating data tailored for specific application domains. These include transforming data to conform to a spatio-temporal grid with minimum distortions and managing complexities such as spatio-temporal autocorrelation issues. Addressing these challenges is pivotal for the effective application of Artificial Intelligence (AI) approaches. Furthermore, adhering to open science principles for data dissemination, reproducibility, visualisation, and reuse is crucial for fostering sustainable research.
Overcoming these challenges offers a substantial opportunity to advance data-driven Earth system research, unlocking the full potential of an integrated, multidimensional view of Earth system processes. This is particularly true when such research is coupled with innovative research paradigms and technological progress.
\end{abstract}

\section{Introduction}

Humanity possesses the capability to observe and model the majority of Earth's subsystems, generating vast amounts of data with unprecedented resolution, quality, and coverage \citep{simmons2016observation, peng2021roadmap, bauer2021digital}. The co-interpretation of these diverse datasets represents an unprecedented opportunity for understanding the intricacies of the Earth system \citep{runge2019causation_earth, mahecha2020earth, tuia2023artificial}. However, this wealth of heterogeneous data comes with substantial challenges. The sheer volume of data, characterised by variations in spatial and temporal resolution as well as data curation levels, coupled with the high complexity of processes encoded in these multi-dimensional datasets, renders conventional data processing and interpretation methods unsuitable \citep{boulton2018challenges, sudmanns2020big}. 

Recognising the need for a simple yet robust data infrastructure to facilitate Earth system data interoperability led to the emergence of various data cube concepts \citep[][and others]{nativi2017view, baumann2019datacubes, giuliani2019paving, kopp2019achieving, mahecha2020earth}. We refer to Earth System Data Cubes (ESDCs) as frameworks where diverse datasets are integrated into a unified, highly interoperable system, organised on a common spatio-temporal grid (a more formal definition is given in Section~\ref{ESDCs}). The essence of ESDCs is to convert the vast array of Earth system data into readily accessible data streams, apt for a variety of Earth system research domains. Such frameworks are gaining widespread acceptance in Earth system research as a solution for managing complex Earth Observation (EO) data.

Given the simplicity of such structures, various initiatives have greatly enhanced the use of EO data derived from satellite remote sensing and other large-scale array data, such as climate model outputs. Initiatives building on an ESDC concept originally developed their data in hand-crafted ways \citep[e.g. ][]{mahecha2020earth, estupinan2021regional, walther2022fluxneteo} or created systems supporting on-demand generations of ESDCs \citep[e.g. ][]{appel2019demand, killough2018open_data_cube, schramm2021openeo}. Earth system data providers have invested tremendous efforts in compiling extensive data catalogues, which can be used for the development of further ESDCs. Notable examples of such catalogues are provided by Google Earth Engine \citep[GEE, ][]{gorelick2017gee}\footnote{\url{https://developers.google.com/earth-engine/datasets/}}, Microsoft Planetary Computer\footnote{\url{https://planetarycomputer.microsoft.com/catalog}}, or the Open Geospatial Data Catalogue of Amazon Web Services (AWS)\footnote{\url{https://aws.amazon.com/earth/}}. Additionally, there is a constant effort to increase the adoption of ESDCs (including generation and analysis) within cloud environments \citep{zellner2024mooc_cubes}. Therefore, ESDCs can be efficiently generated and used in virtual laboratories, such as the  DeepESDL \citep{brandt2023deepesdl,sturm2023deepesdl}\footnote{\url{https://deepesdl.readthedocs.io/}}, or the agricultural virtual lab\footnote{\url{https://agriculturevlab.eu/}}. 
 
This access to straightforward aligned Earth system data has facilitated numerous Earth system research questions. For instance, researchers have employed both linear and non-linear dimensionality reduction methods to generate global indicators for the terrestrial biosphere \citep{kraemer_summarizing_2020}, uncover the main modes of Earth system variables \citep{bueso2020nonlinear_pca}, quantified spatial dynamics of vegetation responses to ENSO in South America \citep{EstupinanSuarez2023Spatial}, or gained major insights on Land Use and Cover Change \citep[LUCC, ][]{santos2019som_lulc_eo_datacubes}. Specifically, EO data cubes, or ESDCs comprising satellite remote sensing imagery, have been instrumental in applications such as learning the vegetation response to climate drivers using Recurrent Neural Network (RNN) architectures \citep{martinuzzi2023learning_biosphere_rnns}, quantifying drought legacy effects on gross primary production \citep{yu2022drought_legacy_hainich}, and detecting spatio-temporal extreme events \citep{mahecha2017detecting}.

However, if the goal is for ESDCs to evolve and become sustainable data infrastructures, it is essential to develop robust ESDC life cycles. Considering the unique characteristics of ESDCs, we cannot merely apply existing research data life-cycle concepts; instead, we must identify and address the peculiarities specific to ESDCs. It is necessary to create opportunities for continuous improvement and to address current challenges by leveraging contemporary technological advancements, specifications, and research paradigms. For instance, data formats and sharing protocols must evolve to align with the current status of cloud-based technologies and standards, in accordance with the adoption of FAIR Open Science principles \citep{wilkinson2016fair}. Moreover, transforming heterogeneous data into an analysis-ready format aligned with a multidimensional spatio-temporal grid is often complex and subject to application-specific variations \citep{giuliani2019paving, zuefle2021mining_eo_cubes}. 

The resulting data format, though straightforward and relatively easy to analyse, encompasses inherent complexities \citep{bejar2023dggs_eo_data_cubes}. These intricacies necessitate careful consideration during data analysis, requiring a profound understanding of the nature of Earth system processes. Naive analyses based on ESDCs can potentially lead to misleading interpretations as pointed out, e.g. by \citet{meyer2018target_oriented, russwurm2023spherical_harmonics_embeddings} or \citet{sweet2023xai_interpretability}. Common pitfalls include model performance inflation caused by spatio-temporal auto-correlation, biased sampling, and inaccurate spatial aggregations.  It's only by adequately addressing these challenges that the full potential of ESDCs can be realised, aligning with the perspectives of various authors \citep{reichstein2019deep, irrgang2021towards,hsieh2022ml_evolution, sun2022review, persello2022deep, tuia2023artificial}. Topics widely discussed today are generative processes in Artificial Intelligence (AI) that could enable researchers to reconstruct unseen data \citep{rttgers2019typhoon, oyama2023gan_super_resolving}. Another promising direction is the potential for making causal inferences solely from data \citep{runge2019causation_earth, krich2021functional, christiansen2022toward, camps2023discovering}. Also, integrating physical constraints and domain knowledge in the inference process can lead to more plausible semi-empirical predictions \citep{ilie2017reverse, karniadakis2021physics, campsvalls2021physics_geosciences, cortes2022physics}. Concurrently, advances in data processing and visualisation technologies not only enhance data exploration and analysis but also aid in disseminating research findings \citep{sochting2023lexcube_ieee}.

This paper seeks to identify the challenges inherent in the complete ESDC life cycle while, at the same time, highlighting the potential to advance Earth system research through these data structures. The manuscript is organised as follows: Section~\ref{art} introduces the concept of ESDC and its relationship to information-preserving systems for Earth system data. In Section~\ref{interoperability}, we elaborate on the ESDC life cycle, displaying the obstacles encountered during data processing and proposing pathways toward creating analysis-ready ESDCs. Section~\ref{datascience} explores the transformative possibilities stemming from contemporary AI advancements in Earth system research while Section~\ref{datascience_challenge} cautions against the risks of uninformed ESDC analysis. Section~\ref{technical} addresses the technical facets of manipulating ESDCs throughout their life cycle, offering insights into technologies that can streamline Earth system data processing. Lastly, in Section~\ref{communication}, we examine the challenges associated with data visualisation in the context of ESDCs. Through this paper, we aim to outline the complexities and opportunities associated with employing ESDCs, hopefully paving the way for advancements in Earth system research.

\section{The Art of Data Cubes}\label{art}

Data cubes are renowned for their capacity to serve as multidimensional arrays of data, enabling the representation of values across various dimensions of interest within a specific domain. Specialised data cubes designed for analytical queries in database systems, such as Online Analytical Processing \citep[OLAP, ][]{chaudhuri1997olap} cubes, have been integrated with Geographical Information System (GIS) databases to give rise to Spatial OLAP \citep[SOLAP, ][]{rivest2005solap} cubes. SOLAP infrastructures, although traditionally associated with vector data, are also available for raster data \citep{kasprzyk2017raster_solap}. Database systems have proven effective in storing and managing Earth system data in the form of data cubes, exemplified by array database solutions like Rasdaman \citep{baumann1998rasdaman}. Additionally, data cube infrastructures can be employed to store indexed files \citep{killough2018open_data_cube}, thus safeguarding the information that might otherwise be lost during data transformation processes, such as reprojection. Here, we rely on a specific interpretation of data cubes, specifically tailored to tackle the vast volumes and interoperability challenges of Earth system data. We first explain the concept of ESDCs, but also provide an overview of related information-preserving structures, namely image collections and information-preserving data cubes, showcasing how they interface with ESDCs.

\subsection{What are Earth System Data Cubes (ESDCs)?}\label{ESDCs}

The concept of ESDCs was introduced along with the Earth System Data Lab \citep[ESDL, ][]{mahecha2020earth}, an integrated data and analytical hub that aimed to unify multiple heterogeneous Earth system data streams into a standard data model with a unique Coordinate Reference System (CRS). ESDCs represent multidimensional data structures designed to facilitate streamlined access, analysis, and manipulation of Earth system data. ESDCs comprise labels as \textbf{dimensions} defining the cube's axes, an array of \textbf{grids} with their associated coordinate values distributed along these dimensions, and univariate \textbf{data} associated with each grid cell. Furthermore, in this paper, we add a new component: a suite of \textbf{attributes} that characterise the data, the dimensions, and the complete ESDC entity.

The \textbf{dimensions} are a set of labels describing the axes of the ESDC. Generally, these dimensions comprise space (e.g. ``x'' and ``y''), time, and variables. Nevertheless, further dimensions can be added (e.g. ``pressure levels'', ``model ensembles'' or ``time series components''). It is crucial to emphasise that while ESDCs conventionally incorporate spatial and temporal dimensions (e.g. latitude, longitude, and time), they are not confined to this paradigm (cf. Table 1 of \citealp{mahecha2020earth}). ESDCs can exhibit different dimensions, and the number of dimensions is called the order of the ESDC. Thus, an increment in the ESDC's complexity according to its dimensions is given by their order (e.g. a spatio-temporal grid of a univariate ESDC has an order of 3, while the order of a multivariate ESDC is 4).

The grouping of \textbf{grids} consists of discrete subsets derived from the domain of each dimension's axis. The values of these subsets are referred to as coordinates, and, in the case of a regular grid, they determine the data's resolution along that specific dimension. For instance, a grid determining a resolution of 0.5 degrees for the ``latitude'' dimension in a global ESDC may have a set of coordinates $\textrm{grid}(\textrm{latitude})=\{-89.75,-89.25,...,89.25,89.75\}$. While coordinates are often associated with numerical values (e.g. latitudes and longitudes), they can encompass a wide range of values. For instance, timestamps in a ``time'' dimension with a set of coordinates $\textrm{grid}(\textrm{time})=\{\textrm{``2022-01-01''},\textrm{``2022-01-02''},...,\textrm{``2022-12-31''}\}$, or components derived from a time series decomposition approach in a ``component'' dimension with a set of coordinates $\textrm{grid}(\textrm{component})=\{\textrm{``raw''},\textrm{``trend''},\textrm{``seasonal''},\textrm{``residual''}\}$. The grids within an ESDC exhibit the following characteristics: 1) In the case of spatial dimensions, they reference the same CRS, 2) the coordinates within a grid share identical units, and 3) they must consist of at least two coordinates; otherwise, the dimension (and consequently the grid) is omitted. It's important to note that, given these properties, irregular grids are also possible, with the temporal dimension grid being a typical example in EO data due to the irregular revisit times of some satellite missions (e.g. Sentinel-2).

The array of \textbf{data} represents scalar values corresponding to each grid cell. Typically, the data spans from observed measurements to modelled values. Nevertheless, one can also encounter higher-order features (i.e. data derived from operations performed on the original values), such as outcomes from time series decomposition or AI-generated products. Furthermore, flag values, which delineate data status, can be incorporated. Cells without data are denoted as ``NA'' (i.e. not available).

The collection of \textbf{attributes} comprises a series of key-value objects that provide additional details about the data. These objects serve as metadata and can offer descriptions ranging from individual variables (including their associated dimensions) to the entire ESDC. The information contained within these attributes typically encompasses a wide range of elements, including, but not limited to, names, acronyms, units, flag definitions, versions, and source details.

\subsection{Relation of ESDCs to Image Collections and Data Cubes}\label{esdc_imagecollections_datacubes}

\begin{figure*}[t]
\begin{center}
    \includegraphics[width=1\textwidth]{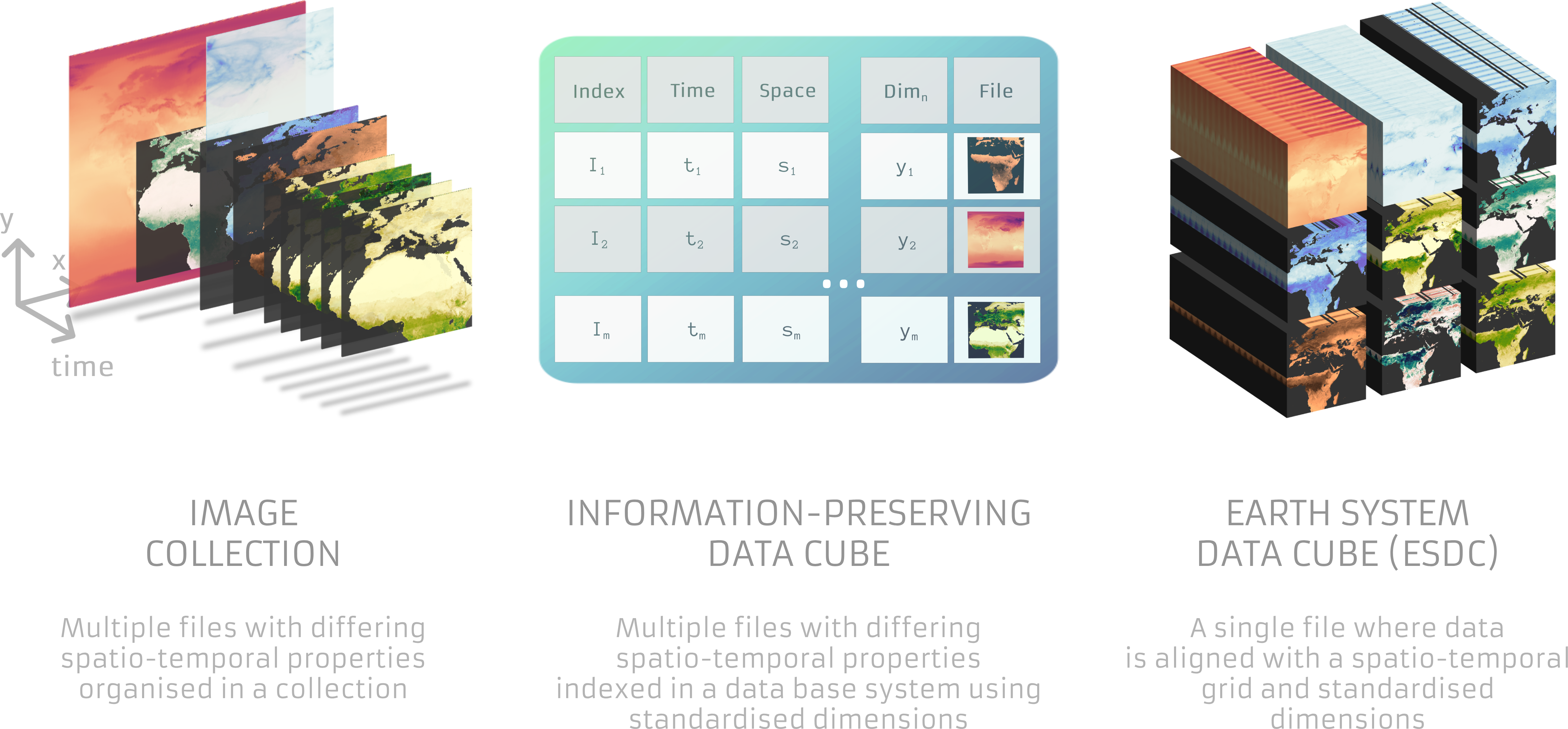}
    \caption{Representations of different storage systems for gridded data in Earth system research: Image collections (left), information-preserving data cubes (centre), and Earth system data cubes (ESDCs, right). Differences in these abstract representations have deep implications for data storage systems, accessibility, interoperability and metadata definitions}
\label{fig:cubes_comparison}
\end{center}
\end{figure*}

Earth system data often exhibits heterogeneity and irregularity, particularly within EO data. Variability can manifest in different spatial resolutions, time units, projections, formats, and more, sometimes even within the same data product. Consequently, two robust approaches to retaining data integrity without succumbing to information loss due to transformations (e.g. reprojection, reduction, and resampling) are to utilise \textbf{image collections} (refer to \citealp[]{appel2019demand} for a comprehensive distinction between image collections and conventional data cubes) or to adopt a process where original files are stored and indexed within a \textbf{information-preserving data cube} infrastructure based on their file metadata (Figure~\ref{fig:cubes_comparison}). In the latter approach, the original files can be stored locally or in the cloud while preserving the essential information intact.

A successful example of \textbf{image collections} is the GEE Catalogue. This extensive, multi-petabyte catalogue stores data in tiled images, where each image may encompass multiple bands, thereby preserving essential information. Furthermore, these images can be organised into an image collection if they share relevancy. GEE also offers the computational resources necessary for accessing and analysing their catalogued data. Within GEE, data cube-like operations can be seamlessly executed through dynamic on-the-fly reprojection, resampling, and reduction for the tiles where a subset of pixels was explicitly requested. It is worth noting, however, that users are required to conform to the specific Application Programming Interfaces (API) provided by GEE for processing and analysing the data effectively.

Standardising image collections and their access brings simplicity and promotes data usage across platforms. Currently, a widely recognised standard is the Spatio-Temporal Assets Catalog (STAC) specification. This specification empowers users to query data assets based on metadata and spatio-temporal criteria. Coupled with domain-specific API clients available for multiple programming languages (cf. Section~\ref{software}) and GIS software (e.g. QGIS STAC Plugin\footnote{\url{https://github.com/stac-utils/qgis-stac-plugin}}), users can easily retrieve data. The flexibility of the STAC specification has prompted numerous data providers to adopt it for creating their own data catalogues\footnote{\url{https://stacindex.org/catalogs/}}, with notable examples including Microsoft Planetary Computer Catalogue\footnote{\url{https://planetarycomputer.microsoft.com/}} and the United States Geological Survey (USGS) Landsat Archive Catalogue (stored in the Amazon Simple Storage Service, S3)\footnote{\url{https://www.usgs.gov/core-science-systems/nli/landsat/landsat-commercial-cloud-data-access}}.

The \textbf{information-preserving data cube} approach is exemplified by the Open Data Cube (ODC) initiative, a prominent model in this field \citep{killough2018open_data_cube,killough2020odcadvances}\footnote{\url{https://www.opendatacube.org/ceos}}. This approach has played a pivotal role in informing governmental actions and policies, as evidenced by their integration into national and regional data cube frameworks \citep{dhu2019national_datacubes,sudmanns2022cube_local}. Noteworthy instances of these initiatives include Digital Earth Africa \citep[DE Africa, formerly known as Africa Regional Data Cube, ][]{killough2019de_africa}, Digital Earth Australia \citep[DE Australia, previously Australian Geoscience Data Cube, ][]{lewis2017australian_datacube,dhu2017de_australia}, the Colombian Data Cube \citep[CDCol, ][]{ariza2017colombian_datacube,bravo2017colombian_data_cube_arc,villamizar2018scaling_cdcol}, and the Swiss Data Cube \citep[SDC, ][]{giuliani2017swiss_datacube}.

While both of these approaches excel in preserving data integrity and offering flexibility for various analyses, achieving Earth system data interoperability necessitates their integration into a unified structure through ESDCs. These ESDCs can be constructed from either approach. For instance, in the case of image collections, it is feasible to request pixels from GEE \citep{nicholas2023gee_pixels}, and data transformations can be executed within the GEE cloud-based environment before downloading the data. It's important to note that limitations related to the size of the requested data can be a potential concern in this process. Alternatively, STAC simplifies the process, particularly when combined with cloud-ready formats. This lazily enables the creation of ESDCs. In the information-preserving data cube approach, platforms like ODC offer a comprehensive system for transforming original data into ESDCs and even provide mechanisms for storing the resulting ESDCs within the data cube infrastructure\footnote{\url{https://www.opendatacube.org/overview}}. Noteworthy is openEO \citep{schramm2021openeo}, an API striving to connect multiple backends, including image collection providers (e.g. GEE) and information-preserving data cubes (e.g. ODC)\footnote{\url{https://openeo.org/software.html}}, to generate ESDCs.

\section{The ESDC Life cycle}\label{interoperability}

\begin{figure*}[t]
\begin{center}
    \includegraphics[width=1\textwidth]{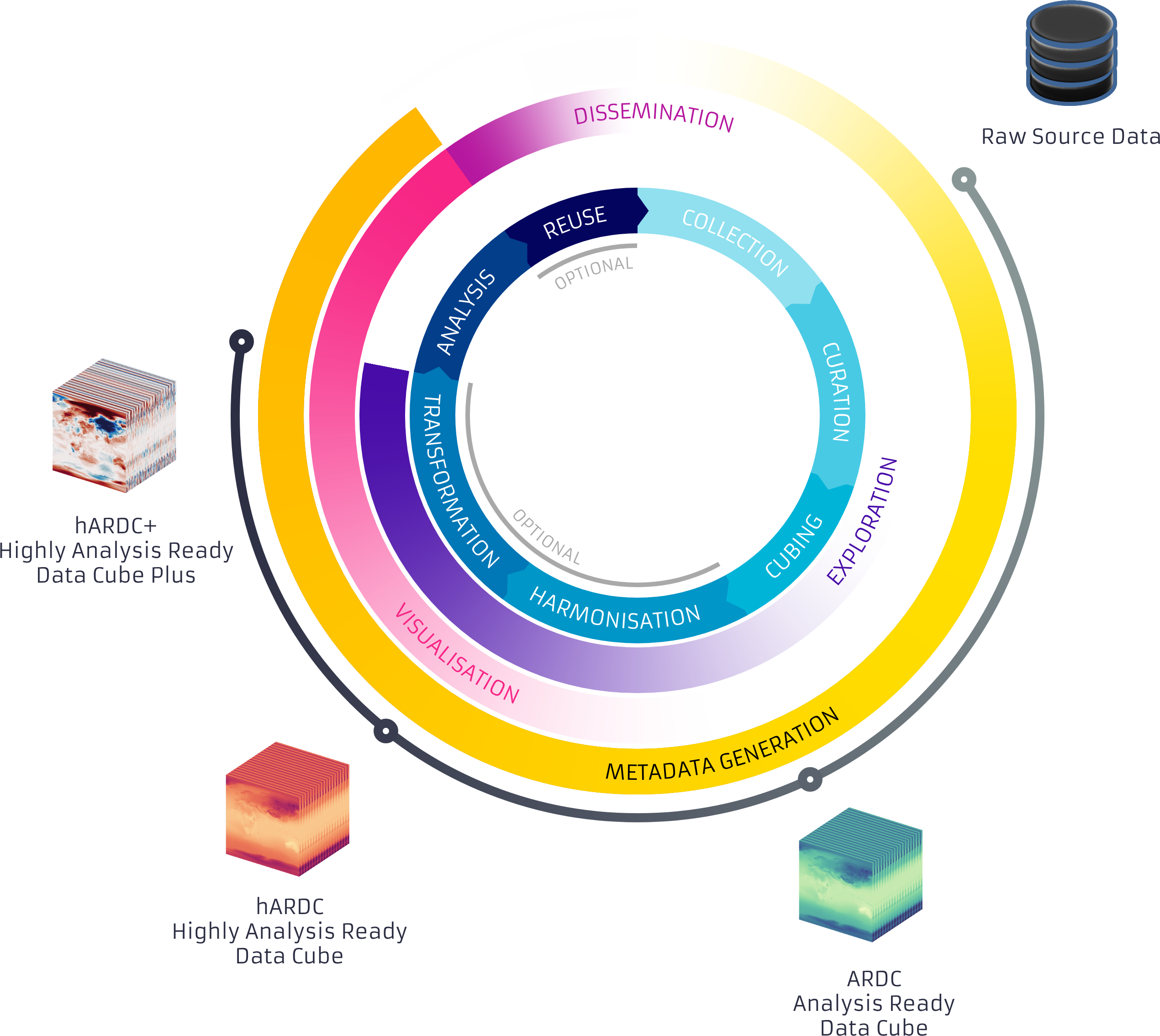}
    \caption{ESDC life cycle. The inner circle represents data processing tasks, and the outer circles represent ancillary tasks that run parallel to the processing steps, involving activities such as data exploration, visualisation, dissemination, and metadata generation. The outermost circle of the diagram illustrates the readiness level of the processed ESDCs at specific points within the cycle}
\label{fig:data_cube_life_cycle}
\end{center}
\end{figure*}

Creating an ESDC from multiple sources, including source files, data cubes, or image collections, is a multifaceted process. The ESDC life cycle, as illustrated in Figure~\ref{fig:data_cube_life_cycle}, encompasses several crucial stages, each playing a vital role in the generation, analysis, and effective utilisation of these data structures. The ESDC life cycle comprises the following key phases: data collection, curation, cubing, harmonisation, transformation, analysis, and reuse. These phases are linked, reflecting the meticulous efforts involved in ESDCs' development. In parallel to these stages, metadata generation occurs concurrently with data transformations, data exploration, visualisation, and dissemination. This section provides an overview of the ESDC life cycle, emphasising relevant considerations that contribute to the streamlined development and utilisation of ESDCs.

\subsection{Collection}\label{collection}

Given that data providers frequently utilise diverse formats and protocols for data sharing, particularly in proportion to the multidimensional complexity of the data, the establishment of streamlined access mechanisms becomes imperative. Traditionally, File Transfer Protocol (FTP) servers have been used for data sharing. However, to enhance data discoverability and usability, data providers are increasingly adopting data stores that offer persistent and standardised data storage. Repositories play a vital role in this process by standardising metadata, enabling easy search and retrieval of assets through metadata queries. Recently, more and more data providers offer APIs to facilitate efficient querying of metadata and access to the data itself by adopting specifications such as STAC and enabling range requests for cloud-optimised data (e.g. Zenodo recently started to support HTTP range requests\footnote{\url{https://blog.zenodo.org/2021/12/07/2021-12-07-hardening-our-service/}}). 

The flexibility of these specifications enhances data interoperability by enabling the development of extensions that simplify data integration. For instance, the Electro-Optical STAC-extension\footnote{\url{https://github.com/stac-extensions/eo}} has been created to facilitate the integration of multispectral remote sensing data by expanding the capabilities of STAC to accommodate specific requirements and metadata associated with this kind of data. Looking ahead, the advantages of data interoperability may potentially extend beyond the realm of raw source data, encompassing entire ESDCs. The datacube STAC-extension\footnote{\url{https://github.com/stac-extensions/datacube}}, currently in a candidate state, has been developed with the primary objective of advancing the integration and interoperability of structured data representations like ESDCs within the STAC ecosystem. This effort aims to broaden the scope of opportunities for reusing ESDCs in new data processing pipelines.

Additionally, the efficiency of data access and collection is contingent upon data formats. GeoTIFF is arguably the most used and renowned data format for georeferenced raster data. This format adds a standard specification for the TIFF format that describes the spatial properties of the raster. It is widely used for EO products such as Landsat imagery. The need to operate in cloud environments has driven the development of cloud-optimised geospatial data formats. Consequently, the GeoTIFF format has evolved to the Cloud-Optimised GeoTIFF (COG)\footnote{\url{https://www.cogeo.org/}} format, enhanced to function efficiently in cloud environments through HTTP range requests. COGs offer several advantages over traditional GeoTIFFs, including reduced latency in data retrieval, faster visualization of large datasets, and a tiled structure that enables parallel processing. The significance of this format is underscored by its recent approval as an Open Geospatial Consortium (OGC) standard\footnote{\url{https://docs.ogc.org/is/21-026/21-026.html}}\textsuperscript{,}\footnote{\url{https://www.ogc.org/press-release/cloud-optimized-geotiff-cog-published-as-official-ogc-standard/}}.

When the dimensionality of the data increases, formats such as NetCDF or HDF5 are typically used to encapsulate data and coordinate values. Tiling and chunking allow efficient access to big data arrays for both data formats. However, these formats are not inherently optimised for cloud environments.The Zarr specification\footnote{\url{https://zarr.dev/}} addresses this limitation and can be used directly in cloud environments, offering several advantages over NetCDF and HDF5. Zarr enables more efficient chunk access for parallel processing, provides better support for distributed computing, and offers improved read and write speeds, particularly in cloud storage systems. Moreover, Zarr's flexible chunking scheme allows for optimised data access patterns, and its simpler metadata structure facilitates easier data discovery and management. Additionally, specifications such as geo-zarr\footnote{\url{https://github.com/zarr-developers/geozarr-spec}} and the xcube dataset convention\footnote{\url{https://xcube.readthedocs.io/en/latest/cubespec.html}} have been introduced to further enhance data interoperability and compatibility within the context of Earth system data.

\subsection{Curation}

Effective data curation stands as a critical anchor in the preparation of data for subsequent spatio-temporal processes and analysis via ESDCs \citep{marujo2022ard_brazil_data_cube}. The transformation of raw data into Analysis-Ready Data (ARD) has emerged as an essential prerequisite across multiple initiatives. ARD ensures that data are readily amenable to queries, analysis, and application development. Notable instances of these initiatives include DE Africa\footnote{\url{https://www.digitalearthafrica.org/platform-resources/analysis-ready-data}}, DE Australia\footnote{\url{https://www.dea.ga.gov.au/about/analysis-ready-data}}, and the Brazil Data Cube \citep{ferreira2020brazil_data_cube,marujo2022ard_brazil_data_cube}, among others.

The Committee on Earth Observation Satellites (CEOS) has precisely defined ARD as ``satellite data that have been processed to a minimum set of requirements and organized into a form that allows immediate analysis with a minimum of additional user effort and interoperability both through time and with other datasets''\footnote{\url{https://ceos.org/ard}}. 
In this definition, ARD also exhibit interoperability both across time and with other datasets (refer to \citealp{siqueira2019card4l} for an overview of the CEOS ARD for Land initiative, CARD4L). CEOS has established a comprehensive set of Product Family Specifications (PFS) tailored to various data groups, including Surface Reflectance, Surface Temperature, Polarimetric Radar, and more. These specifications undergo rigorous peer review processes across multiple satellite platforms, such as Landsat and Sentinel collections, to obtain CEOS ARD approval. It's worth noting that there are ongoing efforts to develop additional PFS, including Interferometric Radar and LiDAR Terrain and Canopy Height. It is also worth noting that the OGC has recently addressed the CEOS ARD concept by forming a new Standards Working Group (SWG) to define a generic multi-part standard specifying a set of minimum requirements for geospatial products to be considered ARD\footnote{\url{https://www.ogc.org/press-release/ogc-forms-new-analysis-ready-data-standards-working-group}}.

It is important to recognise that achieving an ARD level can extend beyond minimum standard specifications. Obtaining ARD often involves crucial preprocessing and data curation tasks that are tailored to the unique requirements of the application domain. For instance, in the context of EO data, these tasks may encompass but are not limited to, cloud and cloud shadow masking (refer to \citealp{skakun2022cmix} for a comprehensive intercomparison exercise of multiple cloud and cloud shadow masking methods), snow masking \citep[e.g. ][]{richiardi2021revised_snow_cover}, and the correction of Bidirectional Reflectance Distribution Function (BRDF) effects to derive Nadir BRDF Adjusted Reflectance (NBAR) values \citep[e.g. ][]{roy2016brdf}.

\subsection{Cubing}\label{cubing}

The concept of ARD may exhibit some subjectivity depending on the specific application. This subjectivity pertains to the data that populates an ESDC. In contrast, ESDCs inherently represent straightforward yet robust analysis-ready integrated entities, capable of simplifying a broad spectrum of analytical tasks \citep{baumann2019datacubes}. An ESDC filled with ARD is often called an Analysis-Ready Data Cube (ARDC), a concept widely employed in DeepESDL. To generate an ARDC, the critical step involves aligning data onto a unified grid. Domain experts predefine this grid, and all data sources must conform. Furthermore, the efficacy of the ARDC processing is significantly influenced by the implementation of an optimal chunking strategy for this grid. This strategy must be tailored to facilitate efficient data processing across diverse analytical scenarios. For instance, analyses focused on temporal dynamics benefit from chunking strategies that preserve the temporal dimension, whereas spatial analyses or cartographic visualisations are optimised by maintaining spatial dimensions within chunks. In scenarios requiring multi-temporal spatial analysis, a hybrid approach combining both temporal and spatial preservation in chunking can be advantageous.

When the grid moves in the spatio-temporal domain, the varying spatio-temporal resolutions and coverage among multiple data sources require selecting adequate methods to fit the data into the predefined grid. Datasets with varying spatial resolutions and coverage must be resampled onto a standard spatial grid. This process often requires modifying the data \citep{cracknell1998pixel}. While non-destructive algorithms such as nearest neighbours can preserve data values (at the cost of duplicating or ignoring values), significant differences in spatial resolution often require transformation through (non-) linear resampling methods, such as cubic convolution or advanced fusion techniques \citep{nikolakopoulos2008fusion}. Complex AI methods can be employed to perform spatial transformations while preserving the quality of the measured variable \cite[e.g. multi-image super-resolution algorithms, ][]{michel2022sen2venus,razzak2023misr}. Another often overlooked issue arises when dealing with extensive variables. In such instances, it is crucial to ensure that, for example, mass balances are not distorted in the process of creating new products.

It is important to note that the application of resampling methods, particularly in the context of generating Global ESDCs covering the entire planet \citep[e.g. ][]{mahecha2020earth}, may introduce geometric distortions. Projecting global datasets onto a plane can distort the data in terms of area, distances, and angles \citep{snyder1989projections}, posing challenges for subsequent analysis (cf. Section \ref{geochallenge}). This can be alleviated by using a Discrete Global Grid System \citep[DGGS, ][]{kmoch2022dggs}. This kind of grid system seeks to minimise distortions, harmonise cell sizes and maintain consistent distances from neighbours. Defining standards and solutions for efficient chunk storage, subsetting, and integration into the ESDC framework will be a challenging future task. Still, it could lead to significant improvements in both the performance and accuracy of spatial algorithms. 

In the case of Regional ESDCs \citep[e.g.][]{estupinan2021regional}, which may cover entire continents, oceans, or administrative regions at various hierarchical levels, selecting an appropriate CRS is crucial to ensure minimal geometric distortion. On local scales, Local ESDCs (also referred to as mini cubes, \citealp{requena2021earthnet}) cover smaller areas of interest \citep[e.g.][]{walther2022fluxneteo}, ideally characterised by high spatial resolutions ranging from sub-meters to meters. Using local ESDCs together with a local CRS enables to minimise distortions.

When dealing with datasets characterised by varying temporal grids, even if they share the same date-time units, irregular temporal grids may emerge. These discrepancies can introduce temporal gaps within the time dimension. In cases where datasets exhibit varying date-time units, especially when working with datasets featuring finer date-time units (e.g. daily records), it becomes necessary to aggregate them to align with a predefined coarser temporal grid (e.g. monthly records). While this process is straightforward for regularly sampled data, it can pose challenges for EO data with long revisit periods (e.g. Landsat data). These challenges can potentially introduce uncertainties during aggregation. Substantial gaps in EO data can have a detrimental impact on the accuracy and representativeness of the aggregated results. This concern is further exacerbated when additional gaps arise due to data disturbances, such as cloud and shadow interference.

\subsection{Harmonisation}\label{harmonisation}

Additional post-processing of data variables may be necessary to address Earth system challenges. This entails further data curation to obtain a fully gap-filled, harmonised product with evenly spaced time steps. In alignment with the naming conventions established for EO data cubes by \citealp{frantz2019force}, we refer to a thoroughly harmonised ESDC as a highly ARDC (hARDC).

Data harmonisation is crucial to ensure the consistency and compatibility of variables obtained or generated using different methodological or technical approaches \citep{wulder2015virtual_constellations}. When discrepancies exist between data measurement or production methods, it can introduce inconsistencies that hinder subsequent analyses involving the specific variables \citep{vogeler2018landlinkr}. To address this, one approach is to create separate variables that represent the same measured quantity, highlighting the differences between them. However, to enhance spatio-temporal resolution and coverage, harmonisation of variables is often necessary \citep[e.g. harmonising reflectance values from Sentinel-2 and Landsat, ][]{claverie2018hls, marujo2023c_factor_landsat_sentinel}.

This can be achieved through simple methods that involve sampling data from the same spatio-temporal index in both variables to establish a direct conversion model \citep[e.g. using matched observations to match Landsat 8 and Sentinel-2, ][]{shang2019landsat_sentinel_tra}. Alternatively, more advanced AI models can harmonise data by incorporating one or more additional variables \citep[e.g. creating a global product of OCO-2 Sun-Induced Fluorescence, SIF, ][]{li2019gosif}. This may require the development of an entire AI pipeline to extend a variable with newly available data or reconstruct it, especially in cases where the variable was not previously measured \citep[e.g. reconstructing SIF from TROPOMI, ][]{chen2022rtsif}. In this sense, data harmonisation also encompasses projecting data in simulated future scenarios \citep[e.g. projecting vegetation dynamics for the rest of the century, ][]{mahowald2016lai_rcp_projected}. In addition, it is crucial to incorporate uncertainty metrics to facilitate accurate and reliable future analysis using the harmonised data variables (cf. Section~\ref{UQ}).

Additionally, to effectively use algorithms that incorporate temporal structures, such as Recurrent Neural Networks \citep[RNNs, ][]{sherstinsky2020rnns}, a regularly spaced and gapless time dimension is usually required. Hence, data from an irregular time dimension should be aggregated or interpolated to fit into a regular temporal grid \citep[e.g. gap-filling Landsat reflectances on a monthly basis, ][]{moreno_martinez2020histarfm}. A suitable predefined temporal resolution must be selected, and data must be gap-filled. Various gap-filling techniques, ranging from simple linear interpolation to more complex AI-based modelling approaches, can be employed to address this \citep[e.g. using Long Short-Term Memory networks, LSTMs, ][]{ren2022lstm_gapfill}. The choice of the gap-filling method depends on factors such as the data's nature, the desired accuracy level, and the specific requirements of the analysis or application.

\subsection{Transformation}

\begin{figure*}[t]
\begin{center}
    \includegraphics[width=0.75\textwidth]{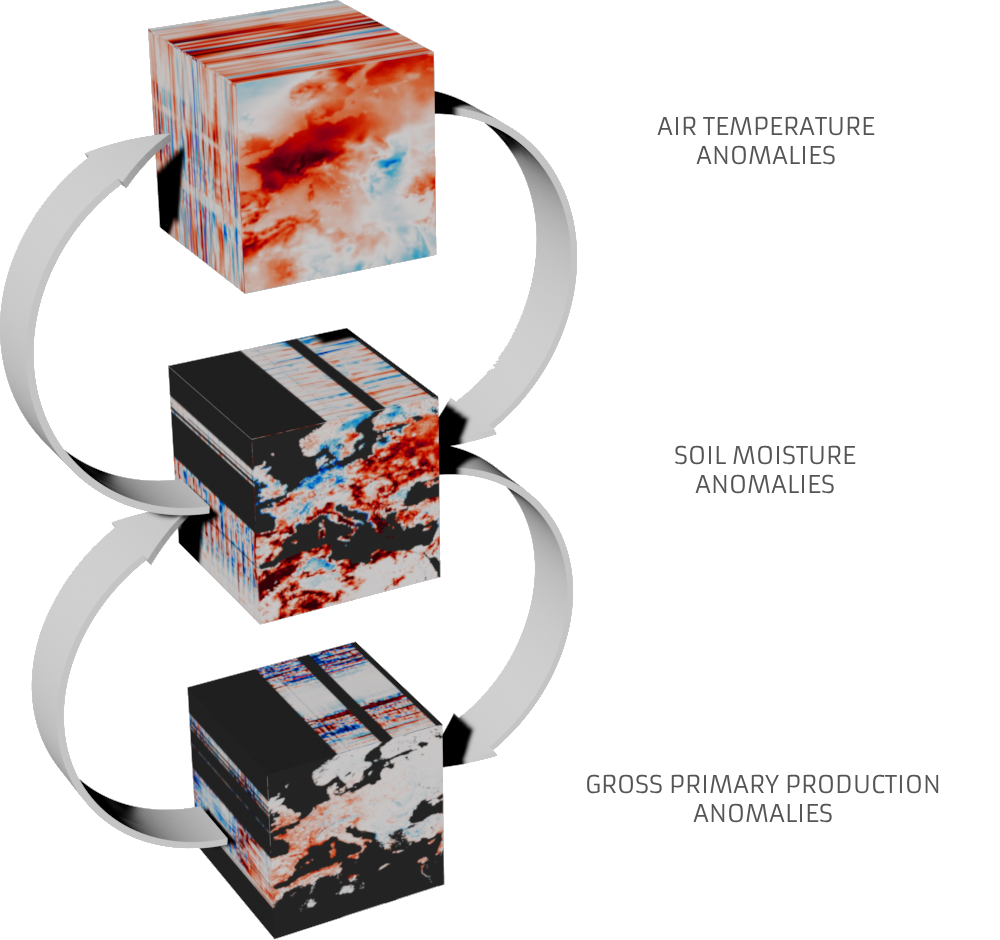}
    \caption{Abstract representation illustrating the connection between three Earth system variables in a hARDC+ (from top to bottom: anomalies in air temperature, soil moisture, and gross primary production). The arrows illustrate the interactions that can be modelled, e.g., predictive modelling (top to bottom) or interpretation (bottom to top), depending on the use case of interest}
\label{fig:anomaly_cubes}
\end{center}
\end{figure*}

Expertly crafted higher-order features often prove highly relevant for addressing Earth system challenges. These new features span a spectrum, encompassing operations that range from simple transformations of the original variables to the creation of entirely novel features derived from advanced AI models. Examples of such features include the computation of spectral indices derived from reflectance bands \citep{montero2023asi}, the extraction of frequencies through time series decomposition \citep{mahecha2010identifying}, the creation of spatio-temporal compositions \citep[e.g. ][]{Griffiths2013bap}, summarising high dimensional dynamics \citep[e.g. ][]{kraemer_summarizing_2020}, and outputs generated by AI models \citep[e.g. ][]{brown2022dynamicworld}. To illustrate, consider a study focusing on climate extremes like heatwaves and droughts' impact on the terrestrial biosphere. In such cases, calculating anomalies for critical variables (e.g. air temperature and soil moisture as proxies for heatwaves and droughts, with Gross Primary Production as the target biosphere variable) is pivotal (see Figure~\ref{fig:anomaly_cubes}). Creating these novel features introduces a new dimension to distinguish between variable values corresponding to raw data and those representing anomalies. In line with the naming conventions introduced by \citealp{frantz2019force}, we designate an ESDC with higher-order features as a hARDC Plus (hARDC+).

\subsection{Reuse}

ESDCs, after generation and analysis, can either evolve into dynamic versions through continuous updates or become static ESDCs, serving as input for the generation of new ESDCs. In the first scenario, establishing a Continuous Integration (CI) pipeline becomes essential for automating ESDC updates. This pipeline can be scheduled to align with the release of new dataset versions, ensuring the ESDC remains current. However, this approach may prove inefficient for EO products that are delivered regularly (e.g. Sentinel-2 or MODIS) and that may be constantly reprocessed by data providers, releasing new versions with updated processing pipelines. In this case, the update schedule should align with the specific needs of the ESDC (e.g. monthly, semi-annually, or annually). In the second scenario, an automatic update of dataset versions is also feasible, eliminating the necessity to extend the ESDC to the most recent date. In either scenario, it is crucial to implement clear reproducibility and traceability practices to ensure the data integrity of future ESDCs.

As highlighted in Section~\ref{collection}, standardisation is pivotal in promoting fluid data interoperability within this context. OGC has recognised the growing significance of data cube approaches for geospatial data. OGC recently established the GeoDataCubes SWG to create an API that facilitates interoperability among various solutions\footnote{\url{https://www.ogc.org/press-release/ogc-forms-new-geodatacube-standards-working-group}}. This standard covers a broad scope, explicitly including API functionalities for access and processing, exchange format recommendations, profiles, and a metadata model.

Additionally, cloud technologies have ushered in the development of data cube services that abstract the underlying file structures and formats, replacing them with APIs offering diverse processing functionalities and promoting interoperability. For instance, platforms like Sentinel Hub\footnote{\url{https://www.sentinel-hub.com/}} serve as sources for ESDC generation through tools like xcube. Moreover, the openEO platform\footnote{\url{https://openeo.cloud}} aims to provide an API that enables connections from multiple clients to various cloud backends using a unified API \citep{schramm2021openeo}. Approaches like these allow the tailored specification of ESDCs on-demand, with server-side processing relieving requesters of the complexities of the generation task. However, this convenience often comes with a trade-off, as the processing engine's code basis, the processing environment, and the input data are not known to requesters. Any modifications to these specifications can result in different outcomes for identical requests to the data cube API, hindering a streamlined update of dynamic ESDCs and a transparent basis for reusing static ESDCs.

In contrast, less convenient but more transparent approaches fully document the ESDC generation process through ``recipes''. These recipes contain versioned source code used for input data processing. Examples include the Pangeo Forge\footnote{\url{https://pangeo-forge.org}} \citep{stern2022pangeo_forge_arco} and DeepESDL recipes\footnote{\url{https://github.com/deepesdl/cube-gen}}. Recipes, coupled with versioned input data and fully specified processing environments, enable practical reproducibility of resulting ESDCs. This approach supports the seamless updating of dynamic ESDCs when new data becomes available and provides transparency for incorporating static ESDCs into new datasets.

Ongoing efforts to enhance data lineage and provenance transparency are integral to the Copernicus Data Space Ecosystem. The development of the ``traceability'' service\footnote{\url{https://dataspace.copernicus.eu/analyse/traceability}}, currently in progress, is designed to empower users to trace all modifications to the data from its origin to its delivery to the end user, ensuring greater transparency and accountability in the ESDC life cycle.

\subsection{Metadata generation}

Traceability and self-explanatory power are essential aspects alongside the data values themselves. When an ESDC is generated, end users may access its description through various sources, including documentation that adheres to best practices for open data publishing within the Earth sciences. Such practices are supported by data journals (e.g. ESSD\footnote{\url{https://www.earth-system-science-data.net/policies/data_policy.html}}) and scientific associations (e.g. AGU Open Science\footnote{\url{https://www.agu.org/-/media/files/publications/your-6-step-guide-for-publishing-open-access-with-agu.pdf/}}), provided that the data producers have furnished comprehensive documentation. However, the data must carry its own encapsulated description in the form of metadata, which typically comprises a set of attributes represented as key-value pairs. This ensures the data contain relevant information about their characteristics, facilitating understanding and utilisation.

Metadata generation should begin at the initial stage of data collection, encompassing crucial information such as data descriptors (e.g. name, units, measurement methods and equipment, resolution), data transformations (e.g. resampling or interpolation methods), metadata transformations (e.g. renaming procedures, conventions conversion), and responsible producers (e.g. creator entity, data provider). This metadata generation process should be consistently maintained throughout the entire ESDC life cycle, documenting each step undertaken to derive the final product. This ensures comprehensive self-contained documentation of the history and processing of the ESDC.

While flexibility exists in metadata management, conventions are crucial when dealing with Earth system data. The Climate and Forecast Metadata Conventions \citep[CF Conventions, ][]{hassell2017cf}, for instance, represent a comprehensive set of standards specifically designed for Earth system data stored in formats such as NetCDF (although they can be readily applied to other formats like Zarr). These conventions facilitate the creation of clear and detailed descriptions of data variables and coordinate dimensions. Furthermore, software like xarray \citep{hoyer2017xarray} can parse CF Conventions and leverage them for different ESDC processes\footnote{\url{https://docs.xarray.dev/en/stable/user-guide/weather-climate.html}}. Compliance with CF Conventions not only simplifies data sharing but also promotes interoperability among various data sources, ensuring that ESDCs adhere to established standards. 

\section{Leveraging ESDCs for Earth system research}\label{datascience}

ESDCs offer promising opportunities for advancing Earth system research, particularly with recent AI developments. This is exemplified for Deep Learning (DL) by the spatio-temporal nature of ESDCs in a tensor-like structure. In this context, several key subjects emerge as highly relevant for Earth system research. We present three pertinent topics where the potential of ESDCs can be leveraged for advancing Earth system research: Physics-Informed Machine Learning (PIML), the adoption of complex sampling strategies, and the quantification of uncertainties.

\subsection{Adding factual knowledge via PIML}\label{piml}

A great addition to Machine Learning (ML) modelling is combining the pure data-driven approach with factual knowledge of the system under investigation \citep{karniadakis2021physics}. PIML leverages domain knowledge (typically mechanistic models or differential equations) and flexible data-driven ML methods (typically neural networks). Consequently, PIML models respect physical boundaries more faithfully while being flexible enough to approximate arbitrarily complex non-linear functions from data 
(cf. discussion and references in \citealp{reichstein2019deep}). 
ESDCs provide a unique structure to access multiple Earth system data streams, and the equation-based model describes the underlying process. Thanks to this ready availability of data and equations, exploring PIML models using a wide array of baseline models would be far easier and faster.
The equations detailing a given variable could be added to the cube as a sub-field of the variable of interest in the same way that space and time are. The eventual implementation should consider the multi-platform and multi-language nature of ESDCs. As illustrated above, this requires a unified and robust approach that suits multiple use cases.

\subsection{Sampling for AI in a complex system}\label{sampling}

Sampling on ESDCs is essential for learning the concrete interactions of drivers, spatial conditions, timing, and other determinants of specific processes and their implications. This involves strategically selecting a manageable subset from the ESDC. This selection process is particularly important for ML algorithms, as they rely on these subsets to establish a foundational understanding of the process to be analysed \citep{atkinson2022sampling, nikparvar2021mlspatial}. Pseudo-random sampling facilitates a broad and diverse data selection, while regionalised sampling uses specific patterns within the data for a more targeted analysis. The latter proves particularly advantageous when the research goal is to comprehend specific phenomena.

\begin{figure*}[t]
\begin{center}
    \includegraphics[width=0.85\textwidth]{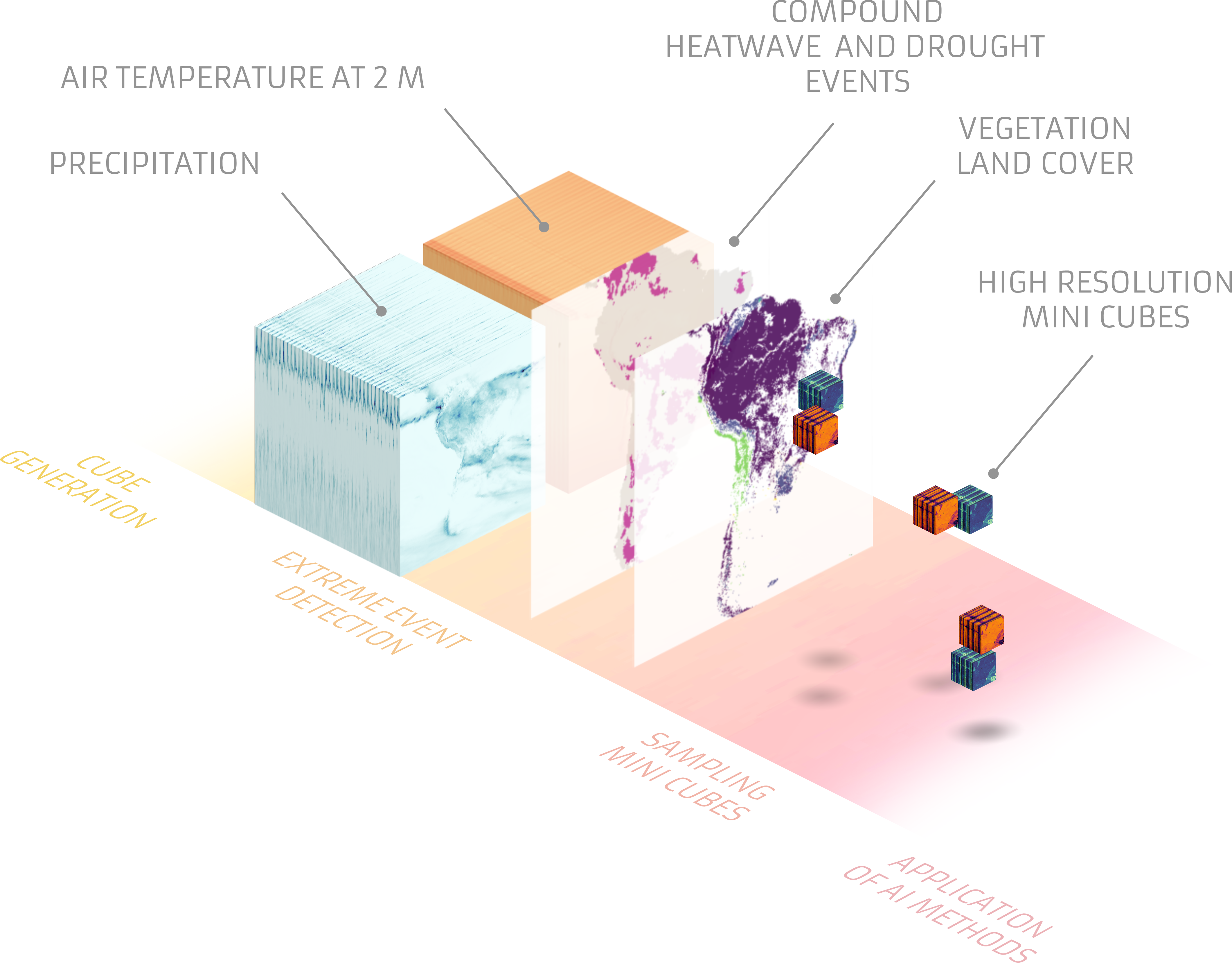}
    \caption{Abstract representation illustrating the process of sampling high-resolution mini cubes for further analysis by considering vegetation land covers and extreme events detected via a global ESDC. Note that sample mini cubes are specified in the spatial and temporal ranges of the detected extreme events (also considering their occurrence)}
\label{fig:sampling}
\end{center}
\end{figure*}

Constructing representative samples in Earth system processes must ensure an unbiased representation of the target variable. The multidimensional nature of Earth system processes poses sampling challenges across multiple variables. Consider, for instance, a study that aims at understanding the effects of climate extremes on the terrestrial biosphere using AI \citep{sippel2018drought}. We know that climate extremes such as heatwaves, droughts, extreme precipitation, flooding, etc., are typically associated with multiple variables \citep{flach2021vegetation}. Additionally, such events can co-occur in unfavourable sequences, i.e., compounding heatwaves, droughts, or floods following droughts \citep{zscheischler2020typology}. To understand such circumstances, one should consider the full spatio-temporal extended in all relevant dimensions, including derived meta-variables that describe the characteristics of these events, such as timing, duration, extent, and intensity \citep{flach2017multivariate}. Often, additional factors gain significance. For example, ecosystem responses to extremes vary in space depending on ecosystem conditions \citep{mahecha2017detecting}, land-cover types \citep{flach2021vegetation}, and associated impacts, e.g., on the carbon cycle \citep{sippel2018drought}. Building suitable AI models that predict such impacts requires including static data (e.g. vegetation type). 

Yet, the critical question is then: how to obtain adequate and balanced training and validation data? Earth system processes often involve rare events of extreme conditions, which may occur sporadically over time and space. This rarity can lead to imbalanced datasets, where certain classes of the target variable or ranges of continuous values are underrepresented. This also applies to ranges of continuous values in an imbalanced distribution. Imbalanced datasets affect the performance and generalisation of models trained on these samples. Achieving spatio-temporal representativeness in this context can be challenging. To train ML algorithms for effective recognition and understanding of these events, it is crucial to include additional sampling within the specific domains where these events occur. For example, when constructing datasets for global flood \citep{li2023glh} or cloud detection \citep{aybar2022cloudsen12}, the methodology involves initiating automatic sampling that covers a broad spectrum of ecosystem conditions. Simultaneously, manually selected events are introduced. This approach ensures a balanced representation of different classes in the dataset, thereby enhancing the algorithm's capability to accurately predict such events. Figure~\ref{fig:sampling} showcases a potential workflow where event detection is performed based on global ESDCs, and samples for high-resolution ML are extracted based on a systematic sampling strategy (e.g. \citealp{ji2024deepextremes}). Here, analysing land cover purity is an option (a relatively homogeneous land cover dominated by a single vegetation type allows for easier comparisons and subsequent analyses), as well as incorporating mixed land covers (which introduces heterogeneity and interactions among land cover types), providing more comprehensive information for model training.

Finally, the selection of samples with the necessary data dimensions must align with the chosen algorithm.  For instance, tabular-based algorithms like tree-based methods require 2-dimensional batches (sample and variable), which are selected as individual points from the spatio-temporal domain. DL methods like Transformers \citep{vaswani_attention_2017} or RNNs, e.g., LSTMs \citep[][]{hochreiter1997lstm}, which consider sequence (or positional) dependencies, require 3-dimensional batches (e.g. sample, timestep, variable) and extract samples usually as subsets of time series from the spatial domain. Convolutional Neural Networks \citep[CNNs, ][]{lecun1989cnn} may be used with 4-dimensional batches (e.g. sample, height, width, variable) by taking spatial subsets or grids from the temporal domain. DL methods accounting for both spatio-temporal dependencies, such as 3DCNNs or Convolutional LSTMs \citep[ConvLSTMs, ][]{shi2015convlstm}, require 5-dimensional batches (e.g. sample, height, width, timestep, variable) and extract samples as subsets of ESDCs.

\subsection{Quantifying uncertainties}\label{UQ}

Uncertainty quantification is crucial to Earth science, providing a comprehensive assessment of the reliability and confidence associated with scientific predictions, model simulations, and observational data. Capturing and modelling uncertainty is a complex task as it arises from various sources such as data limitations, model approximations, and the inherent complexity of Earth system dynamics. 

Uncertainty can be broadly categorised into two types: epistemic uncertainty and aleatoric uncertainty \citep{KIUREGHIAN2009105}. Epistemic uncertainty refers to the model's confidence in its predictions and is related to the choice of model parameters. Techniques such as Bayesian inference or Dropout can estimate epistemic uncertainty \citep{srivastava2014dropout, gal2016dropout}. Bayesian methods assign probability distributions to model parameters, directly quantifying uncertainty. In DL, dropout-based methods create model ensembles by randomly dropping out units during training, providing a measure of uncertainty based on the variability among the ensemble members. While these techniques may not completely capture the underlying uncertainty due to assumptions made during modelling or training, they are practical and can be employed to estimate uncertainty. These methods can be computationally demanding and time-consuming, mainly when applied to real-time applications. However, advancements in cloud platforms and the Monte Carlo (MC)-Dropout technique have enabled reliable uncertainty estimates, even when working with massive amounts of data \citep{MARTINEZFERRER2022113199}. On the other hand, aleatoric uncertainty is associated with the noise or variability present in the data (e.g. data affected by natural variability, measurement errors, or other sources of noise) and cannot be reduced. Instead, it can be identified and quantified as part of the uncertainty characterisation.

ESDCs involving measurements or modelled data can be accompanied by associated uncertainty values. Data assimilation techniques are key in incorporating data into ESDCs while considering the associated uncertainties. Approaches such as Kalman filtering, variational data assimilation, or ensemble-based assimilation can effectively merge different data sources and quantify the resulting uncertainties \citep{MATHIEU20081258}.

\section{Challenges in ESDC analysis}\label{datascience_challenge}

While ESDCs present significant opportunities, it's crucial to approach them with a well-informed strategy to avoid naive applications of analytical methods. In this section, we describe challenges associated with ESDC analysis, focusing on two key issues: addressing geometric distortions (introduced during the cubing process) and spatio-temporal autocorrelation problems.

\begin{figure*}[t]
\begin{center}
    \includegraphics[width=1\textwidth]{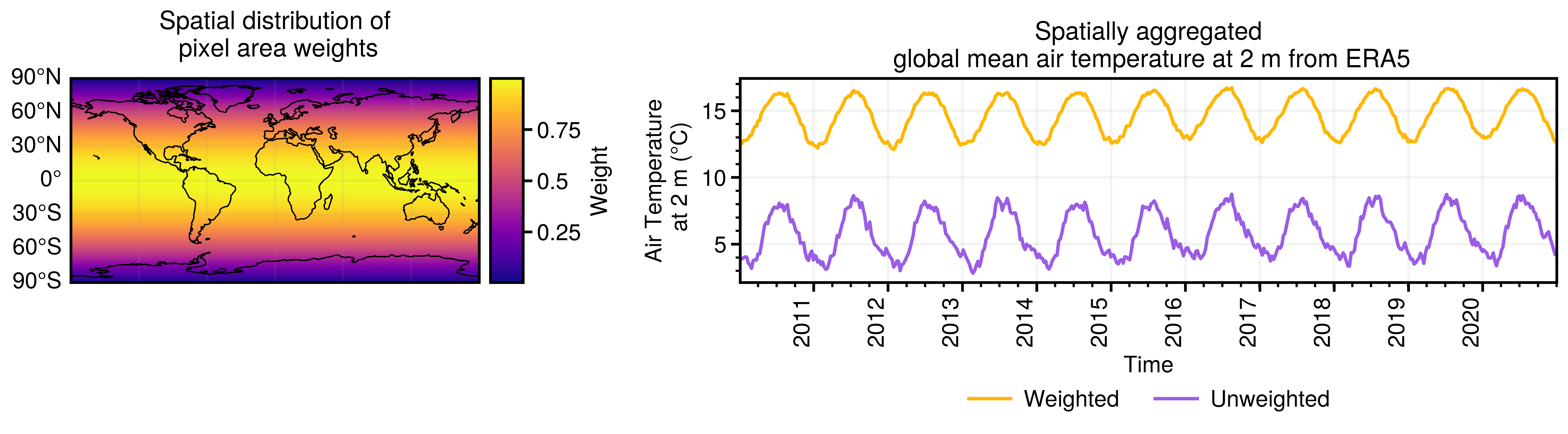}
    \caption{Comparison of air temperature at 2 m from ERA5 with and without weighting on the global mean time series computation. This rather trivial example shows how radically wrong any computation can be if the spherical nature of planet Earth is ignored}
\label{fig:spherical}
\end{center}
\end{figure*}

\subsection{Geometric challenge on planet Earth}\label{geochallenge}

Most ESDCs covering the whole globe use a simple longitude-latitude plate-carr\'ee projection, which fits the ESDC model very well. The approach also allows for efficient storage and subsetting of cubes to user-generated subsets corresponding to a bounding box. However, for advanced data analysis, equirectangular projections have two main drawbacks: 1) grid cells differing in latitude do not have equal area, and 2) the distances to nearest neighbours are not constant.

The first drawback introduces a sampling bias towards high latitudes in the data. This bias can affect the representativeness and accuracy of analyses (cf. Section~\ref{spatiotemporal}), particularly for regions located closer to the equator. The most trivial cases are computations of scalars, like global means (e.g. Figure~\ref{fig:spherical}), which need to be weighted or approaches like principal component analyses that require area-weighed covariance matrices. Effects of this kind have been known for decades and are considered climate textbook knowledge \citep{storch2000statistical}. However, they remain a challenge, as we find them often ignored in ESDC analytics. Issues of this kind can be alleviated using area-weighted statistics, suitable for most linear algorithms, or by performing weighted sampling from grid cells. For advanced, often non-linear data science methods, considering the spherical geometry is much more challenging, and careful consideration is advised before naive applications are performed. Even when applying area-weighted statistics correctly, oversampled areas lead to unnecessary increases in storage requirements and computation time.

The second drawback is particularly significant when applying spatial convolutions or moving window operations. To address this, several approaches can be employed. One option is to use Spherical Harmonics for simple convolutions, providing a transformation that respects the spherical nature of the data \citep{wieczorek2018shtools}. Spherical Harmonics can also be used as coordinate embeddings for neural networks \citep{russwurm2023spherical_harmonics_embeddings}. Another approach involves graph convolutions that consider varying distances to neighbours. 

\subsection{Spatio-temporal representativeness for an accurate model evaluation}\label{spatiotemporal}

Diagnostics on predictive modelling with ESDCs can be challenged by the representativeness and spatio-temporal structure of training data \citep{tobler1970computer,meyer2021predicting, ploton2020spatial, kattenborn2022spatially}. Assessing the accuracy of a prediction is statistically straightforward as long as reference data is available for the entire population or if a respective sample represents the spatio-temporal structure of the population \citep{wadoux2021spatial,brus2021statistical}. However, many modelling tasks build on observations not representative of underlying temporal dynamics or an entire land surface variability (e.g. upscaling functional ecosystem properties from sparse and clustered FLUXNET sites). 
Such an imbalance in reference data may not necessarily lead to a bias in model coefficients \citep{pabon2022potential}. However, it may lead to inflated prediction accuracy estimates, given the commonly limited capacities of ML to extrapolate into the unknown, where the predictor-response relationship may depart \citep{ludwig2023assessing}. Thus, the accuracy assessment of a prediction estimated from clustered samples will not represent the factual accuracy of predictions beyond the reference data availability. This is critical for assessing the quality of a prediction itself and potential error propagation in subsequent analysis \citep{yates2018outstanding, meyer2021predicting, mila2022nearest}. It is advised that predictions should inform on the area of applicability \citep{meyer2021predicting}, i.e., the area in which the predictor-space is covered by the reference data and obtained predictive accuracies thereof are assumed to hold.

However, assessing the predictive performance of a model inside the area of applicability may be challenged by the spatio-temporal structure of the training and test data. Commonly, adjacent observations (both in time and space) are more similar (autocorrelated in space and time), and therefore accuracies determined from test observations near the training data will be more accurate \citep{roberts2017cross, dormann2007methods}. For instance, seasonal effects can inflate model performance when using test observations near training data in the temporal dimension. Dependence among training and reference data results in any case on optimistic estimates of model performance, meaning that such accuracies do not reflect the actual transferability of the model to unseen areas or time steps \citep{roberts2017cross}. For instance, \cite{ploton2020spatial} showed that ML-based models found accurate in the presence of spatial dependent training and validation data may learn spatial data structures instead of transferable relationships between a response (biomass) and the predictors (environmental variables and optical reflectance). This may not only lead to erroneous model transferability and extrapolation to new spatial or temporal domains but also prevent an adequate interpretation of model functioning and attribution to variables and processes \citep{sweet2023xai_interpretability}. Therefore, model performance and interpretation should be performed by minimising spatio-temporal dependence of observations via cross-validation strategies \citep[cf. ][]{roberts2017cross,meyer2018target_oriented,ploton2020spatial,kattenborn2022spatially}.

\section{Technical considerations for managing ESDCs}\label{technical}

Managing ESDCs throughout their entire life cycle is complex and resource-intensive. This section outlines the technical considerations and limitations associated with the current state-of-the-art technological resources for ESDC management. This encompasses aspects such as computing resources, software tools, and scalable solutions that are crucial for effectively handling the challenges involved in ESDC management.

\subsection{Computing resources}\label{computing}

The data size and available computing resources determine data processing feasibility throughout the ESDC life cycle. Computing resources vary from a single laptop to a local cluster with multi-threaded or distributed processing capabilities and can extend to cloud computing environments composed of multiple clusters.
Modern computers are equipped with high-speed Solid-State Drives (SSDs) featuring fast random access and the potential for multiple Gigabytes per second throughput. However, the challenge lies in their limited capacity. In data centres, this is solved by using arrays of disks, but this introduces additional challenges, including latency, throughput, reliability, and security.
Computation on local systems typically involves single-threaded or lightly multi-threaded computations with a higher level of interactivity. In High-Performance Computing (HPC) environments, the software operates in a multi-threaded or multi-core manner and is usually installed by a local system administrator. HPC environments are well-suited for extensive processing tasks but offer reduced interactivity due to the involvement of job schedulers for managing computation resources. Cloud computing environments offer a promising solution for managing vast amounts of Earth system data. These environments can be further improved in terms of scalability by utilising technologies like Kubernetes and Argo, which allow for specialised workflows.  Platforms such as GEE, the European Open Science Cloud (EOSC)\footnote{\url{https://eosc-portal.eu/}}, Google Colaboratory\footnote{\url{https://colab.research.google.com/}}, Amazon SageMaker\footnote{\url{https://aws.amazon.com/sagemaker/}}, DeepESDL\footnote{\url{https://www.earthsystemdatalab.net/}}, and Kaggle\footnote{\url{https://www.kaggle.com/}} provide opportunities for efficient data storage, processing, and collaboration in scientific research. However, it is essential to note that these platforms often impose certain limitations on the users. These limitations include storage capacity, computational resources, available tools for ESDC management, access permissions, and usage restrictions.

\subsection{Software capabilities}\label{software}

In the context of managing ESDCs, diverse tools are available. Here, we present a compendium of useful tools for processing Earth system data within the ESDC life cycle in three prominent programming languages: Python, R, and Julia. 

Python, arguably the most used language for ESDC management, offers {\tt xarray} with labelled multidimensional arrays \citep{hoyer2017xarray}, built on top of {\tt numpy} \citep{harris2020numpy}, and supporting on-disk reading and parallel processing via {\tt dask} \citep{rocklin2015dask} (a Python library for parallel computing, enhancing array objects by employing data partitioning into chunks and employing dynamic task scheduling). Multiple tools are tailored to construct and process {\tt xarray} datasets, which represent ESDCs. For data collection, {\tt rasterio} \citep{gillies2019rasterio}, {\tt rioxarray}\footnote{\url{https://github.com/corteva/rioxarray}}, {\tt satpy} \citep{raspaud2023satpy}, or {\tt EOreader} \citep{maxant2022eoextract_eoreader} are instrumental for reading GeoTIFFs and COGs, returning {\tt xarray} objects. {\tt xarray} excels in reading NetCDF files and cloud-based data via {\tt zarr} as dask-arrays. Vector data can be converted into {\tt xarray} objects using {\tt geocube} \citep{snow2023geocube}. Data sourced from STAC catalogues can be sought through {\tt pystac-client} and directly transformed into {\tt xarray} objects via {\tt stackstac}\footnote{\url{https://github.com/gjoseph92/stackstac}}, {\tt odc-stac}\footnote{\url{https://github.com/opendatacube/odc-stac}}, or {\tt cubo} \citep{montero2024cubo}. These tools support data collection and immediate cubing, including the temporal dimension. GEE enables data retrieval as {\tt numpy} arrays through its API, which can be directly converted into {\tt xarray} objects using {\tt Xee} or {\tt wxee}. GEE's API \citep{gorelick2017gee} and extensions \citep{montero2021eemont} allow data curation before cubing. {\tt xcube} has various data stores for data acquisition and {\tt xarray} object generation\footnote{\url{https://xcube.readthedocs.io/en/latest/plugins.html}}. {\tt XDGGS} \citep{kmoch2024xdggs} simplifies working with different DGGS in {\tt xarray}. The curation, harmonisation, and transformation stages, being subjective and application-dependent, can be accomplished through {\tt xarray} or {\tt numpy} processing. Libraries like {\tt scipy} \citep{virtanen2020scipy}, built on top of {\tt numpy}, offer additional resources leveraging ESDCs as multidimensional arrays.
The analysis phase leverages a plethora of tools. ESDCs as multidimensional arrays are compatible with {\tt numpy}, {\tt scipy}, and related tools. Moreover, ESDCs represented as tensors interface effectively with {\tt tensorflow} \citep{abadi2016tensorflow} or {\tt pytorch} \citep{paszke2019pytorch}. Furthermore, developments that aren't designed for direct ESDC use can also be leveraged using tensors in the representation of ESDCs (e.g. {\tt torchgeo}, \citealp{stewart2022torchgeo}, {\tt GeoTorchAI}, \citealp{chowdhury2022geotorch}, {\tt pytorch-metric-learning}, \citep{Musgrave2020PyTorchML} and {\tt TorchIO}, \citep{prezgarca2021torchio}).

R, a widely used programming language for statistical analysis, has assumed increasing significance in geospatial data processing and management. Raster data sourced from image collections can be managed seamlessly, progressing from data collection to study, with the assistance of libraries like {\tt raster}\footnote{\url{https://github.com/rspatial/raster}} or its more recent counterpart, {\tt terra}\footnote{\url{https://github.com/rspatial/terra}}. ESDCs can be collected and analysed through dedicated tools like {\tt gdalcubes} \citep{appel2019demand} and {\tt stars} \citep{pebesma2023spatial_r_bok}. Regionalised sampling using geospatial data can be conducted using {\tt stpp} \citep{edith2013stpp} and {\tt spatstat} \citep{Baddeley2015spatstat}. Recent developments have introduced the capability for lazy on-disk reading of Zarr files\footnote{\url{https://www.r-bloggers.com/2022/09/reading-zarr-files-with-r-package-stars/}}. Furthermore, data can be sourced and cubed directly from STAC catalogues using {\tt rstac} \citep{simoes2021rstac} in combination with {\tt gdalcubes}. Another comprehensive package for ESDC management is {\tt sits} \citep{simoes2021sits}, offering an end-to-end solution that additionally includes various tools for AI-related tasks, encompassing sampling, tuning, prediction, and the computation of uncertainty values.

Julia, a high-speed programming language, has gained popularity in scientific computing, making it an excellent choice for processing the large volumes of data found in ESDCs. Julia offers tools that cover crucial parts of the ESDC life cycle. These tools include {\tt YAXArrays.jl}\footnote{\url{https://github.com/JuliaDataCubes/YAXArrays.jl}} and {\tt Rasters.jl}\footnote{\url{https://github.com/rafaqz/Rasters.jl}} for multidimensional labelled array operations, {\tt GriddingMachine.jl} \citep{wang2022griddingmachine} for data acquisition, and experimental libraries like {\tt STAC.jl}\footnote{\url{https://github.com/JuliaClimate/STAC.jl}} for data discovery within STAC catalogues. For analysis, Julia provides specialised tools such as {\tt EarthDataLab.jl}\footnote{\url{https://github.com/JuliaDataCubes/EarthDataLab.jl}} for the direct processing of the Earth System Data Cube \citep{mahecha2020earth}. Moreover, data distortions introduced during the cubing process can be addressed using libraries like {\tt OnlineStats.jl}\footnote{\url{https://github.com/joshday/OnlineStats.jl}} \citep{day_onlinestats_2020} and {\tt WeightedOnlineStats.jl}\footnote{\url{https://github.com/gdkrmr/WeightedOnlineStats.jl}} \citep{kraemer_summarizing_2020}. Julia's ecosystem also includes ML tools like {\tt Flux.jl} \citep{innes2018flux}, {\tt DiffEqFlux.jl} \citep{rackauckas2019diffeqflux}, and {\tt ReservoirComputing.jl} \citep{martinuzzi2022reservoir_computing}, enabling advanced data analysis, including novel techniques like PIML.

\subsection{Scalability obstacles}\label{scalability}

The size of ESDCs poses several challenges for analysis. Generally, in most programming languages for data science (e.g. Python, Julia, R), data has to be completely loaded into memory before calculating a simple statistic (e.g. median). However, ESDCs often surpass the memory limit, hindering computations or resulting in significant slowdowns due to frequent disk read-write operations. Instead, users can apply specialised algorithms that calculate statistics iteratively \citep{welford_note_1962,schubert_numerically_2018}.
$\mathcal{O}(1)$ memory algorithms allow the user to track statistics (e.g. mean, sums, and standard deviations) iteratively. They give the user complete control (and responsibility) over the order of the data reads. Because of the spherical nature of the Earth and the resulting differences in the area covered by pixels, these computations require weighted versions of the statistics (cf. Section~\ref{geochallenge}). Errors arising from floating-point arithmetic must be minimised, including the potential for catastrophic cancellation \citep{kahan_further_1965,goldberg_what_1991}.

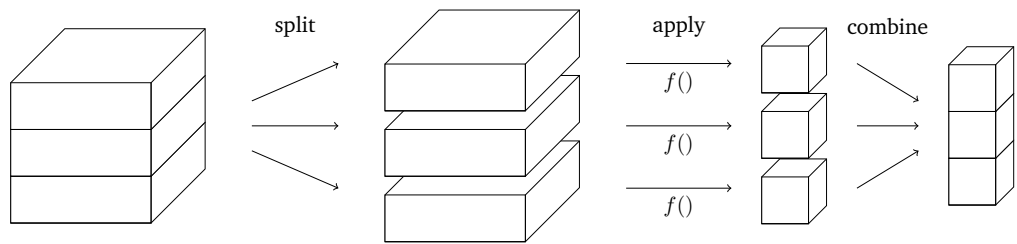
\begin{figure}[t]%
\resizebox{\textwidth}{!}{
\newcommand{\cuboid}[4]{%
  \filldraw[fill = white] #1 -- ++(#2, 0, 0) -- ++(0, 0, -#4) -- ++(  0, #3,   0) -- ++(-#2, 0, 0) -- ++(0, 0, #4) -- cycle;
  \filldraw[fill = white] #1 -- ++(#2, 0, 0) -- ++(0, #3, 0) -- ++(-#2,  0,   0) -- cycle;
  \draw                   #1 -- ++(#2, 0, 0) -- ++(0, #3, 0) -- ++(  0,  0, -#4);%
}
 \begin{tikzpicture}
  \tikzset{
    bigincube/.pic={
      \begin{scope}[shift={(-1.125, -1.125, 1.125)}, xscale = 0.75, yscale = 0.75]
        \cuboid{(0, 0, 0)}{3}{1}{3};
        \cuboid{(0, 1, 0)}{3}{1}{3};
        \cuboid{(0, 2, 0)}{3}{1}{3};
      \end{scope}
    },
    bigoutcube/.pic={
      \begin{scope}[shift={(-0.375, -1.125, 0.375)}, xscale = 0.75, yscale = 0.75]
        \cuboid{(0, 0, 0)}{1}{1}{1};
        \cuboid{(0, 1, 0)}{1}{1}{1};
        \cuboid{(0, 2, 0)}{1}{1}{1};
      \end{scope}
    },
    smallincube/.pic={
      \begin{scope}[shift={(-1.125, -1.425, 1.125)}, xscale = 0.75, yscale = 0.75]
        \cuboid{(0,   0, 0)}{3}{1}{3};
        \cuboid{(0, 1.4, 0)}{3}{1}{3};
        \cuboid{(0, 2.8, 0)}{3}{1}{3};
      \end{scope}
    },
    smalloutcube/.pic={
      \begin{scope}[shift={(-0.375, -1.425, 0.375)}, xscale = 0.75, yscale = 0.75]
        \cuboid{(0,   0, 0)}{1}{1}{1};
        \cuboid{(0, 1.4, 0)}{1}{1}{1};
        \cuboid{(0, 2.8, 0)}{1}{1}{1};
      \end{scope}
    }
  }
  \node[inner xsep=2.3cm, inner ysep=1.8cm] (c)     at (0,  0) {};
  \node[inner xsep=2.3cm, inner ysep=1.8cm] (ci)    at (6,  0) {};
  \node[inner xsep=1cm, inner ysep=1.8cm] (fci)   at (11, 0) {};
  \node[inner xsep=1cm, inner ysep=1.8cm] (fbarc) at (14, 0) {};

  \pic at (c)     {bigincube};
  \pic at (fbarc) {bigoutcube};
  \pic at (ci)    {smallincube};
  \pic at (fci)   {smalloutcube};

   \node at ($ 0.5*(c.east) + 0.5*(ci.west) + (0, 1.6) $) {split};
  \node at ($ 0.5*(ci.east) + 0.5*(fci.west) + (0, 1.6) $) {apply};
  \node at ($ 0.5*(fci.east) + 0.5*(fbarc.west) + (0, 1.6) $) {combine};

  \draw[->] ($ (c.east) + (0,  0.4) $) -- ($ (ci.west) + (0,  1) $);
  \draw[->] (c.east)                   -- (ci.west);
  \draw[->] ($ (c.east) + (0, -0.4) $) -- ($ (ci.west) + (0, -1) $);
  \draw[->] (ci.east)                 -- node[below] {$f()$} (fci.west);
  \draw[->] ($ (ci.east) + (0,  1) $) -- node[below] {$f()$} ($ (fci.west) + (0, 1) $);
  \draw[->] ($ (ci.east) + (0, -1) $) -- node[below] {$f()$} ($ (fci.west) + (0, -1) $);
  \draw[->] (fci)                     -- (fbarc);
  \draw[->] ($ (fci.east) + (0,  1) $) -- ($ (fbarc.west) + (0,  0.4) $);
  \draw[->] ($ (fci.east) + (0, -1)$) --  ($ (fbarc.west) + (0, -0.4)$);
\end{tikzpicture}
}
{
  \caption{ \emph{Split-apply-combine}: \emph{split} an ESDC along arbitrary axes, \emph{apply} a function $f$ to each sub-cube, and then \emph{combine} the results along the same axes that have been used to split the original ESDC}\label{fig:split_apply_combine}
}
\end{figure}

Often, analyses can be performed independently on timesteps, maps, or any other discrete chunks of an ESDC (e.g. dimensions, periods, spatial slices). First, users \emph{split} the data into those chunks, and then \emph{apply} the transformation. In the end, users \emph{combine} the elements back together into a new ESDC (see Figure~\ref{fig:split_apply_combine}). Many analyses can be expressed in terms of \emph{split-apply-combine} \citep{wickham_split-apply-combine_2011,mahecha2020earth}, such as calculating mean seasonal cycle maps from a time axis to a day-of-year axis, or a global mean temperature time series that collapses latitude and longitude into a scalar value per timestep. This method is also known as \emph{map-reduce} in distributed data processing. Still, in contrast, it is made for array-like or tabular data (and the \emph{reduce} step always consists in concatenating the results of the \emph{map} step, cf. \citealp{wickham_split-apply-combine_2011}). Implementations of \emph{split-apply-combine} can trade-off between memory consumption and performance by adjusting the amount of data being loaded into memory simultaneously. They may also take advantage of parallel reading, processing, and writing of data, which is especially important if the data is not stored on local storage but on object stores with high access latency.

Storage in the form of compressed chunks typically employed by ESDCs, where reading a single element requires loading an entire chunk into memory, presents an opportunity for optimising sampling during ML training. Reading points individually is inefficient, as sampling two points from the same chunk necessitates reading the entire chunk twice. To mitigate this, reordering the points within a batch enables reading points from the same chunk jointly, reducing the number of reading operations. Adopting this approach makes it possible to limit the need to read the entire ESDC only once per batch, optimising the data access process.

Ensuring that scalability obstacles are transparent for end users during Earth system data analysis is essential. While experienced users may be able to address scalability issues effectively, less experienced users may struggle with the process if it is not fully transparent. It is important to provide a user-friendly interface that hides the complexities of scalability, allowing users to focus on their analysis tasks. Not all users can access sufficient computing resources for scaling processes, resulting in additional processing costs. Therefore, providing accessible and cost-effective solutions for scalability, such as cloud-based platforms, is crucial to enable a broader range of users to harness the benefits of scaling in Earth system data analysis.

\section{Visual interaction with ESDCs}\label{communication}

Data and process visualisation are critical for communicating Earth system science because big data are often hard to understand intuitively based on metadata alone, especially for non-expert audiences \citep{hibbard2002visualization, kendall2008web, kehrer2012visualization}. The gap between analytic capability and the means to effectively visualise results slows our progress in understanding complex Earth system phenomena. Specialised tools are needed to visualise ESDCs and address their specific needs.  \citet{helbig2017challenges} defined the key challenges of data visualisation for advancing Earth system sciences. Their ambition was to use ESDC visualisation for visual data exploration, facilitating multidisciplinary and collaborative research and also emphasising their educational role. 

\begin{figure*}[t]
\begin{center}
    \includegraphics[width=1\textwidth]{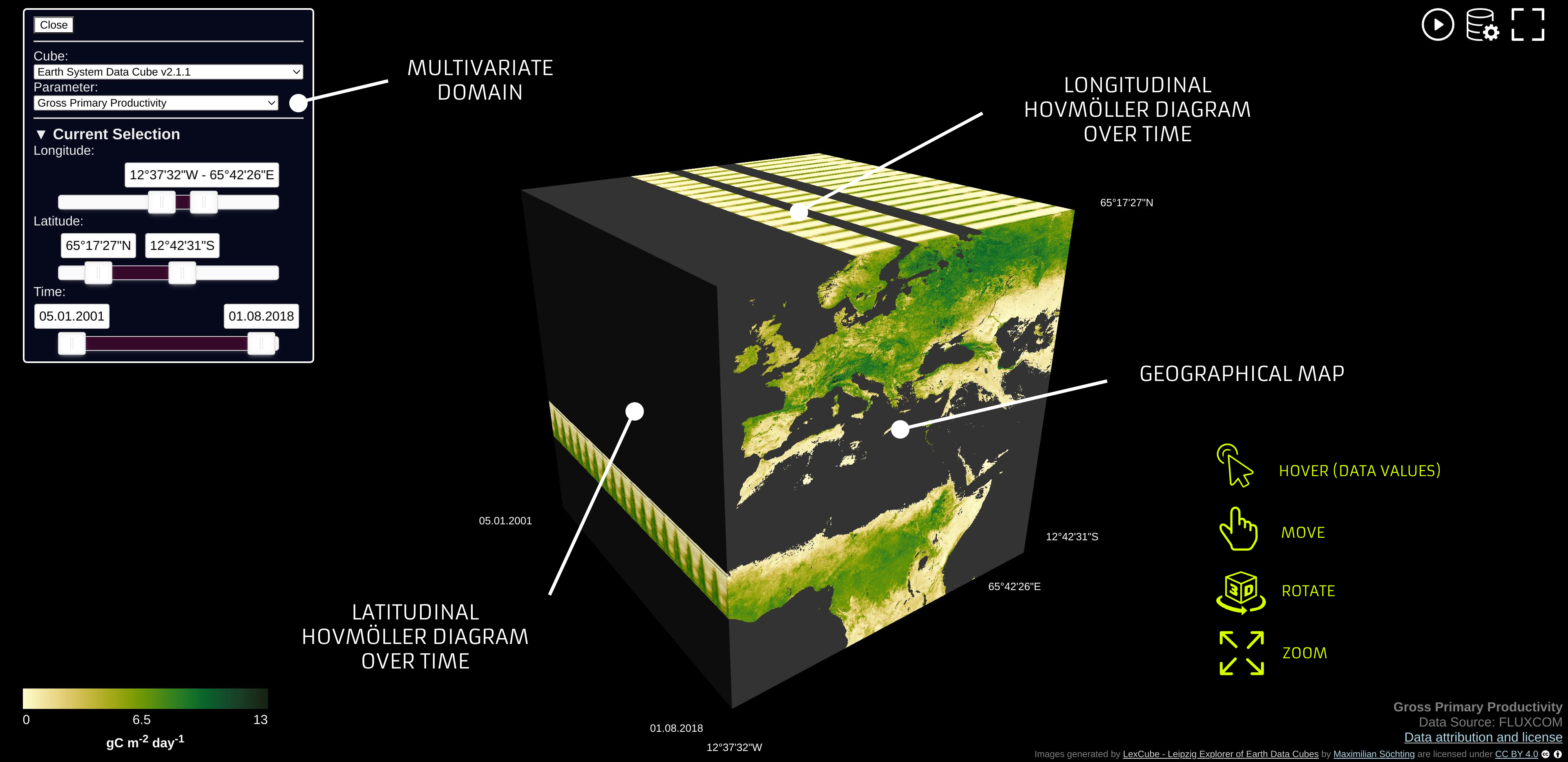}
    \caption{Interactions within an example ESDC in Lexcube, showcasing a geographical map on the front side and Hovm\"oller diagrams depicting temporal changes on the lateral sides. The ESDC allows for interactive subset operations on any side}
\label{fig:vis_interaction}
\end{center}
\end{figure*}

Much progress has been made in visualising ESDCs in Earth system research. Several viewers now have provided researchers with the means to explore and visualise multidimensional environmental datasets and generate scientific illustrations for publications\footnote{\url{https://github.com/carbonplan/maps}}\textsuperscript{,}\footnote{\url{https://cfs.climate.esa.int/}}. However, most approaches still rely on the classical geographical interpretation of georeferenced data and are restricted to displaying maps, extracting singular time series, or Hovm\"oller diagrams. Little advances have been made to visualise ESDCs, particularly multivariate ESDCs, for a better data understanding \citep[cf. static attempts, ][]{mahecha2010identifying, mahecha2017earth, mahecha2020earth}. The long-standing challenge is the trade-off between data interactions not designed for ESDCs and reliance on standard libraries that generate only static visualisations. Recent developments like Lexcube \citep[][cf. Interactions in Figure~\ref{fig:vis_interaction}]{sochting2023lexcube_ieee}\footnote{\url{https://www.lexcube.org/}} and xcube-viewer\footnote{\url{https://github.com/dcs4cop/xcube-viewer}} enable interactive and barrier-free visualisation, allowing users to inspect any ESDC dimension (especially space, time, and variable) interactively. Enabling interactions on large-scale spatio-temporal data on the web is key to democratising our science \citep{steed2014web}.

A significant challenge will be the integration of data analytics with interactive visualisations through visual analytics (cf. the review of \citealp{cui2019visual}). The existing suite of methods is only partially suited for dealing with highly multivariate ESDCs, and most sophisticated visual analytic tools depend on a highly developed local computing infrastructure. There is a pressing need for web-based solutions to address this limitation. The goal should be to incorporate visualisations into any complex workflow to enhance comprehension of data inputs, monitor intermediate outcomes, and observe spatiotemporally structured results. One approach could be the tight integration of visualisation in developer workflows, particularly in popular environments like Jupyter Notebooks. 

Integrating analytics tools with visualisation frameworks would allow researchers to dynamically explore, analyse, and visualise ESDCs in a unified environment in real-time. This would empower researchers to gain immediate insights into the relationships and patterns within the data. Additionally, incorporating visualisation into developer workflows would facilitate seamless visualisation generation at any stage of the ESDC life cycle, allowing researchers to visualise intermediate and final results and facilitating a more intuitive, iterative exploration of Earth system data.

ESDC visualisation extends its potential beyond the scientific community to engage and inform a wider audience. Nevertheless, this is particularly effective when accompanied by expert guidance such as tutorials, workshops, or annotations. Interactive open-access visualisations, exemplified by tools like Lexcube, allow political stakeholders and the general public to directly access and examine climate data (e.g. global or regional climate anomalies and trends). Open-access interactive visualisations enable scientifically literate individuals and those with less technical expertise to delve into ESDCs easily and rapidly by visualising anomalies, trends, and the interplay of variables. 
Such accessibility encourages a broader understanding and appreciation of Earth system research among diverse stakeholders, fostering a more informed and constructive dialogue about climate-related issues.

\section{Conclusions and perspective}

This paper reviews and explores the challenges and opportunities of leveraging ESDCs for Earth system research. This becomes particularly important in developing Earth Digital Twins (i.e. ``a digital replication of the state and temporal evolution of the Earth system'', \citealp{bauer2021dtwin}). In this sense, the topics discussed here are of significance in initiatives like Destination Earth (DestinE)\footnote{\url{https://digital-strategy.ec.europa.eu/en/policies/destination-earth}}.
The inherent simplicity and versatility of ESDCs enable a comprehensive exploration of the complex Earth system, facilitating a deeper understanding of intricate processes and phenomena. For advancing our understanding of the Earth system, the following key considerations emerge and need to be addressed by the research community to tap into the full potential of ESDCs: 

\begin{enumerate}\justifying

    \item \textbf{Artificial Intelligence on ESDCs:} 
    The abundance of large-scale Earth system data, coupled with recent advancements in AI methods, compels the application of the latest developments in deep learning to ESDCs.
    Capitalising on the tensor-like structure of ESDCs in DL and incorporating factual knowledge through Physics-Informed Machine Learning approaches promise great advances in modelling and understanding. Recent advancements in AI, particularly in attention mechanisms, have opened up new possibilities for Earth system research. Techniques such as LLMs, generative image models \citep[e.g. Stable Diffusion, ][]{rombach2021diffusion}, as well as recent image and video segmentation models \citep[e.g. Segment Anything Model, SAM and SAM~2, ][]{kirillov2023sam,ravi2024sam2}, may hold the potential to significantly advance our understanding of the Earth system \citep{wu2023samgeo,osco2023sam4rs}. The ability to `communicate' to ESDCs to extract valuable insights \citep[e.g. ][]{lobry2020visual_question_rs} is within reach (e.g. using text prompts to extract variable anomalies from a specific land cover over a particular region). Furthermore, there is potential to generate ESDCs using text prompts, images, videos, or additional data inputs simultaneously by leveraging the power of multi-modal mechanisms \citep[e.g. ImageBind, ][]{girdhar2023imagebind}, e.g., simulating the impact on vegetation due to an extreme event over a real ESDC using text prompts and geographical data. 
    However, caution must be exercised when applying AI methods to ESDCs to avoid erroneous predictions and interpretations. Factors such as spatio-temporal auto-correlation, the spherical nature of the Earth, and biased sampling in the spatio-temporal and multivariate domains pose risks. Still, the abstract nature of ESDCs provides an opportunity to establish a de facto standard for AI in Earth system science, benefiting from optimised data access and technical enhancements. To ensure reliable outcomes, standardised methods are needed to address spatial dependency, the model's area of applicability, and model uncertainty within ESDC structures.
 
    \item \textbf{Interacting with ESDCs:} 
    The heterogeneity, size, and multivariate nature of datasets also may imply that using ESDCs' is unintuitive, which hampers interpretation. Effective communication opportunities with such data are crucial throughout the ESDC life cycle, both for scientists and a wider audience. Visualisation plays a key role in this regard. While visualisation tools are available to support the analysis process and scientific dissemination, there is still considerable potential for further exploration and development of visualisations. We believe that interactive visualisations are one key, as demonstrated by Lexcube. One promising avenue is the integration of visualisation directly into the analytics workflow (e.g. within Jupyter Notebooks or similar environments), and another is enabling visual analytics of ESDCs. In both cases, the challenge is making such interactions possible during the analysis process to enable the scientific exploitation of large ESDCs.
    
    \item \textbf{Technical challenges of large ESDCs:} 
    The multidimensional nature, varying spatio-temporal scales and resolutions, and applicability of ESDCs imply a series of technical challenges. These include interoperability issues, different geographical projections, interpolation and aggregation questions, and varying readiness levels for further analyses. Ensuring data integrity and interpretability while making Earth system data analysis-ready and interoperable requires tracing and encoding all data transformations and modifications in ESDC metadata. To address these challenges, developing guidelines and standards for geospatial datacubes is crucial for promoting FAIR and Open Earth System Science.
    The ever-increasing size and complexity of datasets demand scalable solutions to tackle associated challenges. The ongoing efforts of the open-source software community are commendable in this regard, as they contribute to the advancement of tools and frameworks tailored to handle big  Earth system data. Furthermore, cloud environments present a possible solution to quickly scale workloads when processing data within the ESDC life cycle. They offer the advantages of on-demand resource allocation and scalability, allowing researchers to access the necessary computational power and storage capacity when needed.
    
    \item \textbf{Integrating (geospatial) data beyond cubes:} 
    ESDCs already offer the potential for advancing Earth system research and analysis in multiple domains. However, ESDCs can benefit from integrating different methodological approaches or data sources at different scales. One example is the integration of Unoccupied Aerial Vehicle (UAV)- and Light Detection and Ranging (LiDAR)-based data. This data provides a means to collect highly localised and high-resolution measurements, making them particularly suitable for localised studies and gaining valuable insights into fine-scale processes. Another example is the integration of vector data\footnote{\url{https://r-spatial.org/r/2022/09/12/vdc.html}}, which typically represents categorical information and carries great importance in multiple Earth system spheres (e.g. socioeconomic features). Additionally, in-situ collections of any process (e.g. via ecological monitoring data) are essential. Today, the quest is that users request the integration of any additional data sources while remaining fully valid. Yet, it poses a challenge as it raises important questions regarding interoperability and the encapsulation of multi-resolution cubes that incorporate multi-scale raster data and the combination of raster and vector data within a unified framework.
    
    \item \textbf{Towards flexible cube-based structures:} 
    To advance ESDCs' benefits, it is essential to advance the standards of ESDC structures and start considering hierarchical data structures, including ESDCs as ``leaves'' (e.g. {\tt xarray}'s DataTree structure) or even unstructured grid systems (e.g. Project Raijin\footnote{\url{https://raijin.ucar.edu/}} with {\tt uxarray}\footnote{\url{https://github.com/UXARRAY/uxarray}}). Given the abundance of insightful (but heterogeneous) datasets, this would enhance Earth system research, regardless of their resolution or dimensionality. Nevertheless, this implies that we must ensure data traceability and interpretability as heterogeneity increases in the resolution or dimensionality domains. A prime example lies in integrating AI models' predictions within ESDCs. In such instances, additional dimensions must be incorporated to capture uncertainties (or quality flag systems) associated with AI-based predictions. This provides valuable insights into the reliability and robustness of the data. Leveraging the power of ESDCs in diverse fields can drive innovation, advance scientific knowledge, and enable more informed decision-making in a wide range of domains.
    
\end{enumerate}

\section*{Acknowledgments}
We are grateful for the European Space Agency (ESA) funding for the DeepESDL and the DeepExtremes projects. Also, we thank the DLR for funding the ML4Earth and VW for funding the Digital Forest project. We also thank the DFG for supporting NFDI4Earth and NFDI4Biodiversity. We thank Pablo Mahecha for generating Figure~\ref{fig:anomaly_cubes} using inputs from Lexcube. Comments by the anonymous reviewers greatly improved the quality of the paper.

\section*{Funding Statement}
This research was supported by grants from the European Space Agency ESA (``AI4Science - Deep Extremes'' and ``DeepESDL''). D.M. and M.D.M. acknowledge support from the ``Digital Forest'' project, Ministry of Lower-Saxony for Science and Culture (MWK) via the program Nieders\"achsisches Vorab (ZN 3679), and the ``RS4BEF'' project via the iDiv's Flexpool program. M.D.M. and M.R. acknowledges support by the German Aerospace Center, DLR representing the Bundesministerium für Wirtschaft und Klimaschutz (ML4Earth, 50EE2201B). M.D.M., M.S., F.C. and F.G. acknowledge support by the Deutsche Forschungsgemeinschaft (DFG, German Research Foundation) for funding the ``NFDI4Earth'', project number: 460036893. T.K. acknowledges support by the Deutsche Forschungsgemeinschaft (DFG, German Research Foundation) for funding the ``PANOPS'', project number: 504978936. C.A. acknowledges support by the National Council of Science, Technology, and Technological Innovation (CONCYTEC, Peru) through the ``PROYECTOS DE INVESTIGACIÓN BÁSICA – 2023-01'' program with contract number PE501083135-2023-PROCIENCIA. M.D.M. and F.M. acknowledge support by the Federal Ministry of Education and Research of Germany and by Sächsische Staatsministerium für Wissenschaft, Kultur und Tourismus in the programme Center of Excellence for AI-research ``Center for Scalable Data Analytics and Artificial Intelligence Dresden/Leipzig'', project identification number: ScaDS.AI.

\bibliographystyle{abbrvnat}
\bibliography{Bib}

\end{document}